\pdfoutput=1

\documentclass[11pt]{article}

\usepackage[]{emnlp2021}

\usepackage{times}
\usepackage{latexsym}

\usepackage[T1]{fontenc}

\usepackage[utf8]{inputenc}

\usepackage{microtype}

\usepackage{CJK}
\usepackage{tabularx}
\usepackage{graphicx}
\usepackage{booktabs}
\usepackage{amsmath}
\usepackage{algorithm}
\usepackage{algorithmic}
\usepackage{float}
\usepackage{color}
\usepackage{arydshln}
\usepackage{bm}
\usepackage{amssymb}
\usepackage{multirow}
\usepackage{multicol}
\usepackage{bbding}
\usepackage{arydshln}
\usepackage{makecell}

\title{CSAGN: Conversational Structure Aware Graph Network for Conversational~Semantic~Role~Labeling}

\author{Han Wu$^{1,2}$, Kun Xu$^{3}$, Linqi Song$^{1,2}\thanks{~~Corresponding author.}$\\
         $^{1}$City University of Hong Kong Shenzhen Research Institute \\
         $^{2}$Department of Computer Science, City University of Hong Kong\\
	 $^{3}$Tencent AI Lab\\
	 \texttt{hanwu32-c@my.cityu.edu.hk}\\
	 \texttt{kxkunxu@tencent.com}\\
	 \texttt{linqi.song@cityu.edu.hk}
}

\begin{document}
\maketitle
\begin{CJK*}{UTF8}{gbsn}
\begin{abstract}
Conversational semantic role labeling (CSRL) is believed to be a crucial step towards dialogue understanding.
However, it remains a major challenge for existing CSRL parser to handle conversational structural information.
In this paper, we present a simple and effective architecture for CSRL which aims to address this problem.
Our model is based on a conversational structure-aware graph network which explicitly encodes the speaker dependent information.
We also propose a multi-task learning method to further improve the model.
Experimental results on benchmark datasets show that our model with our proposed training objectives significantly outperforms previous baselines.
\end{abstract}

\section{Introduction}
\begin{figure}[t!]
    \centering
    \includegraphics[width=1.0\linewidth]{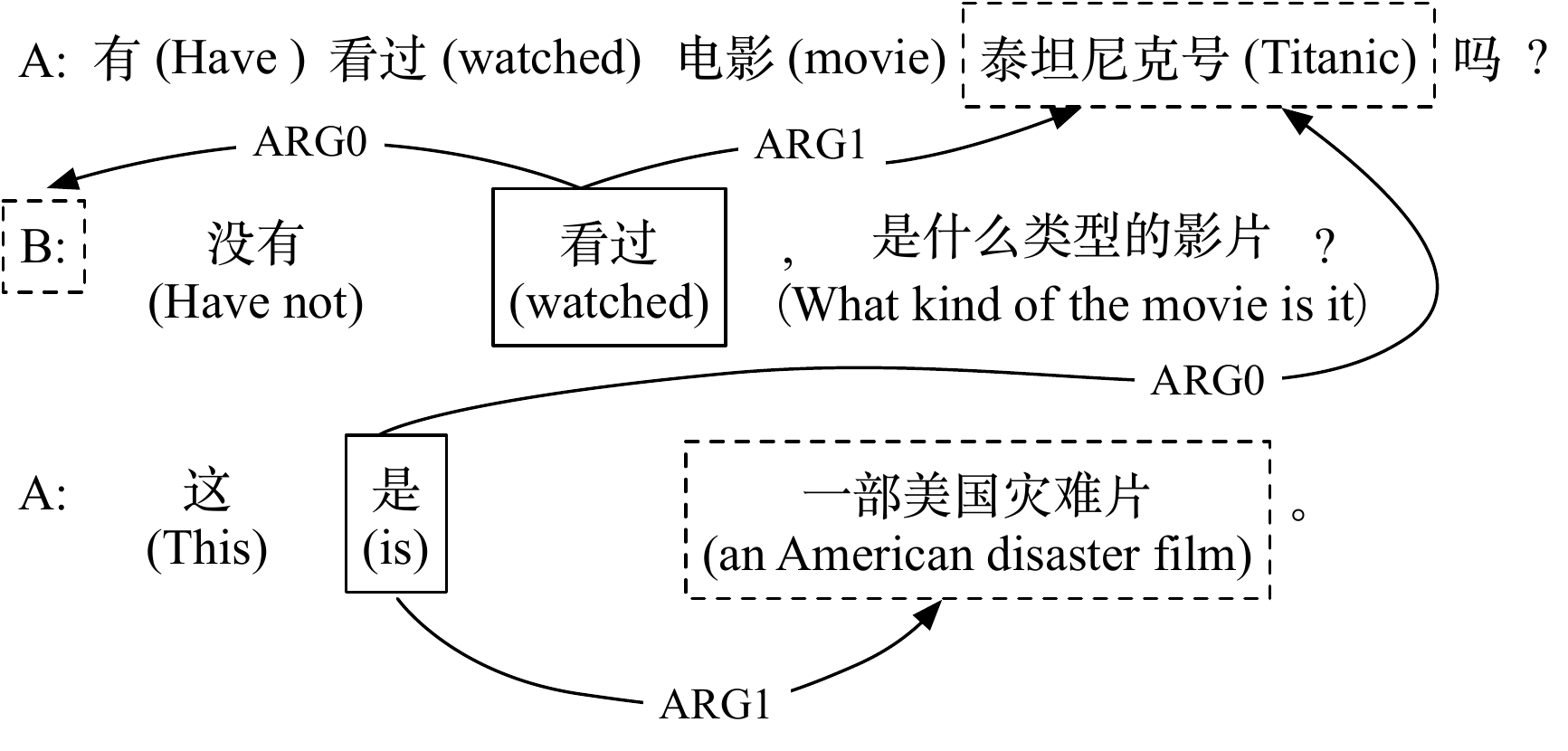}
    \caption{A conversational SRL example.}
    \label{fig:example}
    \vspace{-3mm}
\end{figure}

Recent research has achieved impressive improvements on conversation-based tasks, such as
dialogue response generation \cite{li2017dailydialog,dinan2018wizard,wu2019proactive}, task-oriented dialogue modeling \cite{mrkvsic2017neural,budzianowski2018multiwoz} and conversational question answering \cite{choi2018quac,reddy2019coqa}.
However, the frequent occurrences of ellipsis and anaphora in human conversations still create huge challenges for dialogue understanding.
To address this, \citet{xu2021conversational} proposed the Conversational Semantic Role Labeling (CSRL) task whose goal is to extract predicate-argument structures across the entire conversation. 
Figure~\ref{fig:example} illustrates an example,
where a CSRL parser needs to identify ``{\small《泰坦尼克号》}(Titanic)'' as the \texttt{ARG1} argument of the predicate ``{\small 看过} (watched)" and the \texttt{ARG0} argument of the predicate ``{\small 是} (is)".
One can see that in the original conversation, ``{\small《泰坦尼克号》}(Titanic)'' is omitted in the second turn and referred as ``{\small 这} (this)" in the last turn.
\newcite{xu2021conversational} has demonstrated the usefulness of CSRL on many downstream tasks such as dialogue generation and dialogue rewriting.

Despite the successes that CSRL has achieved, existing CSRL model \cite{xu2021conversational} is 
a simple extension of BERT. Specifically, they first encode each utterance into \textbf{local} contextual representations with pre-trained language models, and then utilize a stack of self-attention layers to obtain \textbf{global} contextual representations.
We argue that their model may suffer two main problems.
First, in the local feature extraction phase, they ignore the fact that jointly considering the predicate and context utterances could help the model better identify some relevant ommited arguments.
Second, in the global feature extraction phase, some vital conversational structural information, such as the speaker information, is not properly encoded in their model.
Indeed, speaker-dependent information is necessary for modelling inter-speaker and intra-speaker dependency,
both of which could help the model to better handle coreference resolution and zero pronoun resolution.

\begin{figure*}[ht!]
    \centering
    \includegraphics[width=1.0\linewidth]{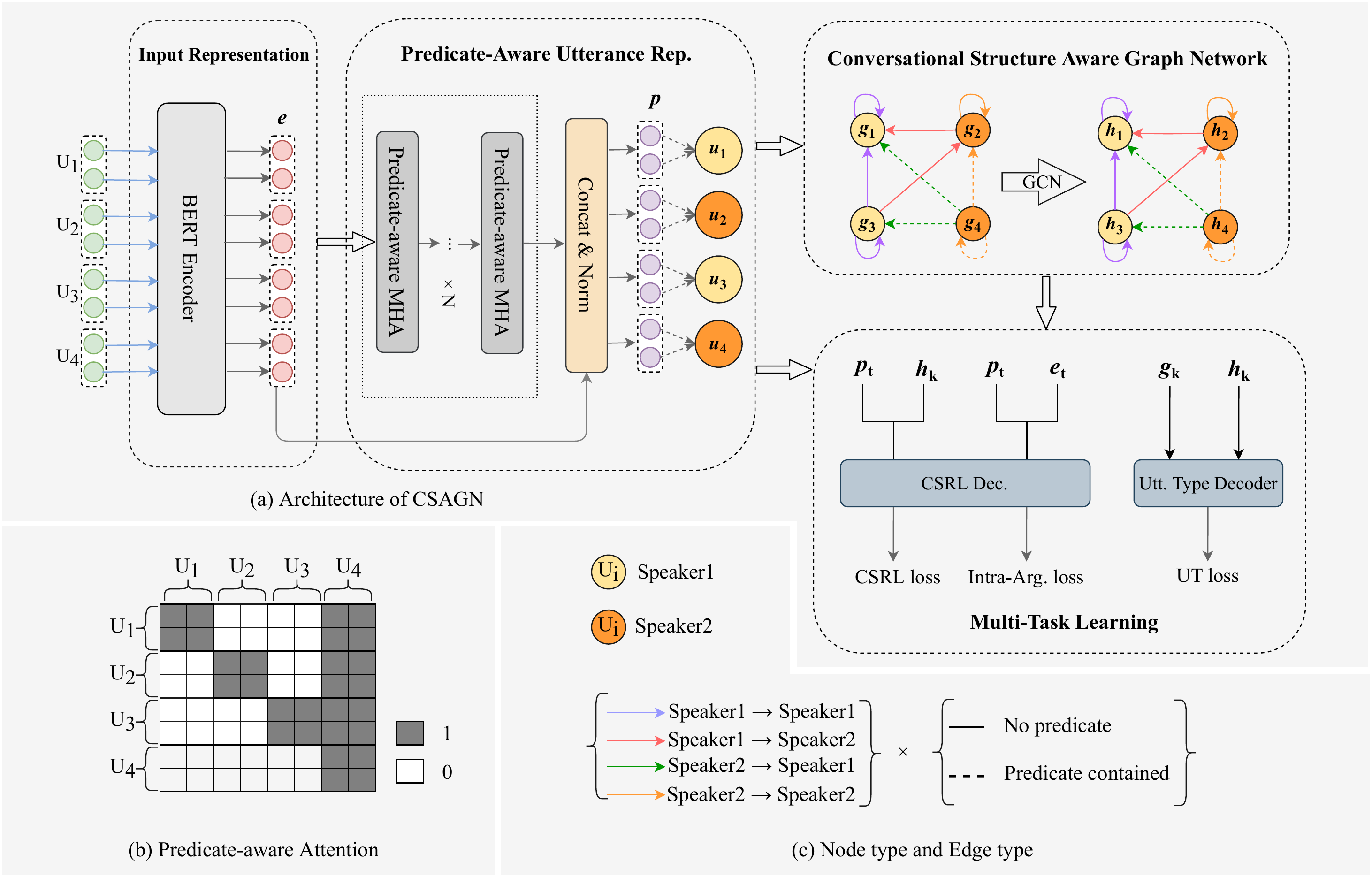}
    \caption{The overview of our proposed CSAGN.}
    \label{fig:model}
\end{figure*}

Motivated by the above observations, we propose a new CSRL model, which consists of three main components.
First, we use a pre-trained language model to generate local contextual representations for tokens (Sec.~\ref{input}), which is similar to \newcite{xu2021conversational}.
Then, we propose a new attention strategy to learn predicate-aware contextual representations for tokens (Sec.~\ref{utterance_representation}).
Finally, we propose a Conversational Structure Aware Graph Network (CSAGN) for learning high-level structural  features to represent utterances (Sec.~\ref{CSAGN}).
The resulted utterance representations are incorporated with token representations obtained in the previous two components.
With the enhanced token representations, our model predicts the arguments for the given predicate.

In addition, we introduce a multi-task learning method with two new objectives.
Experimental results on benchmark datasets show that our model substantially outperforms existing baselines.
Our proposed training objectives could also help the model to better learn predicate-aware token representations and structure-aware utterance representations. Our code is publicly available at \url{https://github.com/hahahawu/CSAGN}.

\section{Model}

The overall architecture of CSAGN is illustrated in Figure \ref{fig:model} which consists of three main components, and we introduce them as follows:

\subsection{Input Representation}
\label{input}
Given a dialogue $\mathbf{C} = (u_1, u_2, ..., u_K)$ of $K$ utterances, where $u_k = (w_{k,1}, w_{k,2}, ..., w_{k, |u_k|})$ consists of a sequence of words, we first use a pre-trained language model such as BERT to obtain the initial context representation $\bm{e}$.

\subsection{Predicate-Aware Utterance Representation}
\label{utterance_representation}

We propose a new attention strategy to better learn predicate-aware context representations for tokens.
Specifically, tokens are only allowed to attend to tokens in the same utterance or the utterance that includes the predicate:

\begin{equation*}
    \textit{Mask}[i,j] = \left\{ 
    \begin{array}{rcl}
        ~1&, & if~\mathbb{U}_i = \mathbb{U}_j~or~\mathbb{U}_j = \mathbb{U}_{pred}\\
        0&, & \textit{otherwise}
    \end{array}
    \right.
\end{equation*}
where $i$, $j$  are the token indexes in the dialogue, $\mathbb{U}_i$ and $\mathbb{U}_j$ are the utterances that the $i$-th and $j$-th token belongs to, $\mathbb{U}_{pred}$ is the utterance that includes the given predicate. For example, in Figure~\ref{fig:model}(b), assuming $U_4$ includes the predicate, previous utterances $U_{1}$, $U_{2}$ and $U_{3}$ could attend to themselves and $U_4$, while $U_4$ only attends to itself.

In practice, we use additional four self-attention blocks \citep{vaswani2017attention} with our proposed attention strategy to learn predicate-aware context representations from $\bm{e}$, which results in token representations $\bm{p}$.
Then, we obtain utterance representations $\bm{u}$ by max-pooling over words in the same utterance.

\subsection{Conversational Structure Aware Graph Network}
\label{CSAGN}
We present the Conversational Structure Aware Graph Network (CSAGN) to capture speaker dependent contextual information in a conversation.
In particular, we design a \textit{\textbf{directed}} graph from the encoded utterances to capture the interaction between the speakers.

Formally, a conversation having $K$ utterances is represented as a directed graph $\mathcal{G} = (\mathcal{V}, \mathcal{E}, \mathcal{R}, \mathcal{W})$, with vertices $v_i \in \mathcal{V}$; labeled edges (relations) $e_{ij} \in \mathcal{E}$ with label $r_{ij} \in \mathcal{R}$ are the relations between vertices $v_i$ and $v_j$; $\alpha_{ij} \in \mathcal{W}$ is the weight of the relational edge $r_{ij}$ with $0 \leq \alpha_{ij} \leq 1$.
Furthermore, the graph is constructed from the utterances in the following way:
\paragraph{Vertices:} Each utterance in the conversation is
    represented as a vertex $v_{i} \in \mathcal{V}$ in $\mathcal{G}$. Each vertex $v_{i}$ is initialized with its corresponding representation in $\bm{u}$, say $g_i$.
\paragraph{Edges:}
    Each utterance (vertex) is contextually dependent on its past utterances in a conversation, thus each vertex $v_{i}$ has an edge with the vertices that represent the past utterances: \{$v_{0}$, $v_{1}$,...,$v_{i-1}$\}.
\paragraph{Edge weights:} We calculate edge weights as follows: for vertex $v_j$, the weight of incoming edge $r_{ij}$ is:
\begin{gather*}
    \alpha_{ij} = \text{softmax}(g_j^T\mathbf{W}_e[g_{0},...,g_{i-1}])
\end{gather*}
where $\mathbf{W}_e$ is the attention matrix learnt from the training.
This ensures that, vertex $v_i$ which has incoming edges with vertices vertex $v_0$,...,$v_{i-1}$ receives a total weight contribution of 1.

\paragraph{Relations:} The relation $r$ of an edge $r_{ij}$ depends upon two aspects:

\textbf{\textit{Speaker dependency}} -- The relation depends on both the speakers of the constituting vertices: $p_{s}(v_{i})$ (speaker of $u_{i}$) and $p_{s}(v_{j})$ (speaker of $u_{j}$).

\textbf{\textit{Predicate dependency}} -- The relation also depends upon whether the utterance $u_i$ or $u_j$ includes the predicate.

If there are $M$ distinct speakers involving in a conversation, then the number of relational edge types is $M$(from\_speaker) * $M$(to\_speaker) * 2 (containing predicate or not) = $2M^2$.

\paragraph{Graph feature transformation} We now discuss how to propagate global information among the nodes. Following previous work \citep{schlichtkrull2018modeling, ghosal2019dialoguegcn}, we use a two-step graph convolution process which essentially can be understood as a special case of messenge passing method \citep{gilmer2017neural} to encode the nodes. We formulate the process as following:

\begin{gather*}
    h_i^{(1)} = \sigma(\sum_{r \in \mathcal{R}} \sum_{j \in \mathcal{N}_i^r} \frac{\alpha_{ij}}{c_{i,r}} W_r^{(1)}g_j + \alpha_{ii}W_0^{(1)}g_i), \\
    h_i^{(2)} = \sigma(\sum_{j \in N_i^r} W^{(2)} h_j^{(1)} + W_0^{(2)} h_i^{(1)}),\\
    for~~i = 1,2,..,K.
    \vspace{-5mm}
\end{gather*}
 \noindent where $h_i^{(l)}$ is the $i$-th encoded node feature from $l$-th layer. $\mathcal{N}_i^r$ denotes the neighboring nodes of $i$-th node under relation $r \in \mathcal{R}$. $\sigma$ is ReLU \citep{nair2010rectified} function. $W_r^{(1)}, W^{(2)}, W_0^{(l)}, c_{i,r}$ are learnable parameters.

After the message propagation, the node representations are updated with the initial node embeddings and the message representations. The final utterance representations are denoted as $\bm{h}$.

\section{Multi-Task Learning}
In this section, we describe how to train the model, based on the representations $\bm{e}$, $\bm{p}$, $\bm{g}$ and $\bm{h}$.

\paragraph{SRL Objective.}
Formally, given an input conversation $\bm{x}$, this objective is to minimize the negative log likelihood of the corresponding correct label sequence $\bm{y}$.
Our model predict the corresponding label $y_t$ based on the token representation $\bm{p_t}$ and its corresponding utterance representation $\bm{h_k}$:
\begin{equation}
\begin{aligned}
    p(y_t|\bm{x};\bm{\theta}) & = p(y_t|\bm{p_{t}},\bm{h_{k}};\bm{\theta}) \\
    & = \text{softmax}(\bm{W_c}[\bm{p_t}\oplus\bm{h_k}])^{T}\delta_{y_t}
\end{aligned}
\end{equation}
where $\bm{W_c}$ is the softmax matrix and $\delta_{y_t}$ is Kronecker delta with a dimension for each output symbol, so $\text{softmax}(\bm{W_c}[\bm{h_t}\oplus\bm{g_k}])^{T}\delta_{y_t}$ is exactly the $y_t$'th element of the distribution defined by the softmax.

\begin{table*}[ht!]
\small
\fontsize{9}{10} \selectfont
\setlength\tabcolsep{4.5pt}
\centering
\bgroup
\def\arraystretch{1.2}
\begin{tabular}{lcccccccccccc}
\toprule[0.8pt]
\multirow{2}{*}{Method} & \multicolumn{3}{c}{DuConv} & & \multicolumn{3}{c}{NewsDialog} & & \multicolumn{3}{c}{PersonalDialog} \\
\cline{2-4} \cline{6-8} \cline{10-12}
& F1$_{all}$ & F1$_{cross}$ & F1$_{intra}$ & & F1$_{all}$ & F1$_{cross}$ & F1$_{intra}$ & & F1$_{all}$ & F1$_{cross}$ & F1$_{intra}$ \\
\hline
SimpleBERT & 86.54 & 81.62 & 87.02 & & 77.68 & 51.47 & 80.99 & & 66.53 & 30.48 & 70.00\\
CSRL-BERT & 88.46 & 81.94 & 89.46 & & 78.77 & 51.01 & 82.48 & & 68.46 & 32.56 & 72.02\\
$\textbf{Ours}$ & \textbf{89.47} & \textbf{84.57} & \textbf{90.15} & & \textbf{80.86} & \textbf{55.54} & \textbf{84.24} & & \textbf{71.82} & \textbf{36.89} & \textbf{75.46}\\
\quad w/o Predicate-Aware Rep. & 88.34 & 82.09 & 89.36 & & 79.94 & 56.46 & 82.87 & & 69.51 & 28.25 & 73.36\\
\quad w/o SAGN & 88.64 & 83.03 & 89.36 & & 77.97 & 50.16 & 81.86 & & 70.32 & 32.41 & 72.24\\
\hline
w/ SRL Objective & 89.10 & 83.88 & 89.73 & & 78.70 & 54.25 & 82.14 & & 69.54 & 29.00 & 73.49\\
w/o Intra-Argument Objective & 88.67 & 83.90 & 89.33 & & 80.55 & 55.87 & 84.00 & & 70.29 & 34.04 & 73.91\\
w/o Utterance-Type Objective & 89.16 & 83.66 & 90.15 & & 79.36 & 54.02 & 82.80 & & 70.70 & 33.04 & 74.81\\
\hline
w/ Full Attention & 88.75 & 83.08 & 89.68 & & 79.09 & 51.78 & 82.64 & & 70.07 & 32.32 & 73.66\\
w/o speaker dependency & 89.03 & 83.84 & 89.56 & & 79.75 & 56.31 & 82.75 & & 70.61 & 34.89 & 74.21\\
w/o predicate dependency & 89.04 & 84.12 & 89.55 & & 79.70 & 54.76 & 83.11 & & 71.06 & 36.23 & 74.66\\
\toprule[0.8pt]
\end{tabular}
\egroup
\caption{Evaluation results on the DuConv, PersonalDialog and NewsDialog datasets.}
\label{tab:results}
\end{table*}

\paragraph{Intra-Argument Objective.}
The arguments in CSRL could be categorized into two classes, i.e., the intra- and cross-arguments,
where the former are in the same dialogue turns as the predicate while the latter usually occur in the dialogue history.
Intuitively, the model should be able to identify the intra-arguments without using the dialogue contextual information. Motivated by this observation, we introduce a new loss function that only uses $\bm{e}$ and $\bm{p}$ to predict the \textbf{intra}-arguments:
\begin{equation}
\begin{aligned}
    \mathcal{L}_{intra} & = - \sum_{t=1}^{n} \log p(y_t|\bm{x};\bm{\theta}) \sigma(y_t) \\
    \mathbf{P} & = [\bm{p_t}, |\bm{p_t} - \bm{e_t}|,\bm{p_t} \odot \bm{e_t}, \bm{e_t}] \\
    p(y_t|\bm{x};\bm{\theta}) & = p(y_t|\bm{e_{t}},\bm{p_{t}};\bm{\theta}) \\
    & = \text{softmax}(\bm{W_c}\mathbf{P})^{T}\delta_{y_t}
\end{aligned}
\end{equation}
where $\sigma(y_t)$ is a boolean scalar that indicates whether $y_t$ is an intra-argument token,
$\bm{W_c}$ is the softmax matrix that used in the SRL Objective.

\paragraph{Utterance Type Objective.}
We additionally introduce a utterance type objective to learn better utterance representations.
Specifically, we classify all utterances into three categories, namely \textit{predicate-utterance} (utterances containing the predicate), \textit{argument-utterance} (utterances containing arguments but not containing the predicate) and \textit{irrelevant-utterance} (utterances without any arguments).
We use utterance representations $\bm{g}$ and $\bm{h}$ to classify the utterance type:
\begin{equation}
\begin{aligned}
    \mathcal{L}_{ut} & = - \sum_{k=1}^{K} \log p(y_k|\bm{g_k}, \bm{h_k};\bm{\theta})
\end{aligned}
\end{equation}

where $y_k$ is the utterance type and $K$ is the total number of utterances.

Finally, we jointly consider these three types of loss: $\mathcal{L} = \alpha_1*\mathcal{L}_{SRL} + \alpha_2*\mathcal{L}_{intra} + \alpha_3*\mathcal{L}_{ut}$,
where $\alpha_1, \alpha_2, \alpha_3$ are hyper parameters.

\section{Experiments}
We evaluate our model on three datasets, i.e., DuConv \citep{wu2019proactive}, NewsDialog \citep{wang2021naturalconv} and PersonalDialog \citep{wu2019proactive}.\footnote{The CSRL annotations of these datasets are provided by \citet{xu2021conversational}}
DuConv is a multi-turn dialogue dataset that focuses on the domain of movie and star,
while NewsDialog and PersonalDialog are open-domain dialogue datasets.\footnote{More details about the annotations on these datasets are listed in Appendix.}
We use the same train/dev/test split as \newcite{xu2021conversational}: DuConv annotations are splitted into 80\%/10\%/10\% as train/dev/in-domain test set while the NewsDialog and PersonalDialog annotations are treated as the out-domain test set.

The hyper-parameters used in our model are listed as follows.
The hop size and embedding dimension of CSAGN is set to 4 and 100, respectively.
The $\alpha_1, \alpha_2, \alpha_3$ are set to 1.0, 1.0 and 1.0, respectively.
The batch size is set to 128. 

\paragraph{Results and Discussion.}
We used the micro-average F1 scores over the (predicate, argument, label) tuples.
Following \newcite{xu2021conversational}, we also evaluate F1 scores over intra- and cross-arguments.
We compare with two baselines that use different strategies to encode the dialogue history and speaker information. In particular, \textit{SimpleBERT} \cite{shi2019simple} uses the BERT as their backbone
and simply concatenates the entire dialogue history with the predicate;
\textit{CSRL-BERT} \cite{xu2021conversational} also uses the BERT but attempts to encode the dialogue structural information by integrating the dialogue turn and speaker embeddings in the input embedding layer.
Table~\ref{tab:results} summarizes the results of our model and these baselines.

We can see that our model significantly outperforms existing baselines on both the in-domain and out-domain datasets.
We can also see that our model benefits from the multi-task training.
In particular, when only using the SRL objective, the F1$_{all}$ scores drop by 0.37, 2.16 and 2.28 on three datasets.
Without using either the intra-argument or utterance-type objective, the performances on all datasets also decrease.
Moreover, we observe that introducing the intra-argument objective could consistently improve F1$_{intra}$ while the utterance-type objective is more important to the F1$_{cross}$.
We think this is because (1) the intra-argument objective denoises the noise from other irrelevant information within the dialogue context; (2) identifying cross-arguments requires a better understanding of the global dialogue structures.

Let us first look at the impact of different components on our model.
From Table~\ref{tab:results}, we can see that both the predicate-aware representation and speaker-aware graph network (SAGN) could improve the F1$_{cross}$ performance. These results indicate that (1) the predicate-aware attention strategy could help the model to better capture the long-distance dependencies between arguments and predicates;
(2) the speaker information encoded in the SAGN is also helpful to identify arguments across dialogue turns.
Furthermore, we also experiment with the full attention strategy to obtain predicate-aware representations, that is, each token attends to all tokens in the entire dialogue.
From Table~\ref{tab:results}, we can see that this strategy achieves worse performance over all metrics.
This result is expected since the full attention treats all utterances equally,
therefore it may encode irrelevant and noisy dependencies into the contextual representation.

Recall that,  we model two types of dependencies in our graph neural network, i.e., the speaker and predicate dependency. We investigate the impact of each dependency on our model and results are shown in Table~\ref{tab:results}.
We can see that removing either dependency may hurt the performance, especially the F1$_{cross}$ score.
This result suggests that these structural information is useful for identifying the cross arguments.

\section{Conclusion}
In this paper, we propose a conversational structure aware graph network for the task of conversational semantic role labeling and a multi-task learning method.
Experimental results on the benchmark dataset show that our method significantly outperforms previous baselines and achieves the state-of-the-art performance.

\section*{Acknowledgement}
We would like to thank the anonymous reviewers for their valuable and constructive comments. This work was supported in part by the Technological Breakthrough Project of Science, Technology and Innovation Commission of Shenzhen Municipality under Grants JSGG20201102162000001, the Guangdong Basic and Applied Basic Research Foundation under Key Project 2019B1515120032, and the City University of Hong Kong Teaching Development Grants 6000755.

\bibliography{emnlp2021}
\bibliographystyle{acl_natbib}


\newpage

\appendix

\begin{table}[t!]
\small
    \fontsize{10}{11} \selectfont
    \setlength\tabcolsep{2.1pt}
    \centering
    \bgroup
    \def\arraystretch{1.2}
    \begin{tabular}{l|cccc}
    \toprule
        Dataset & \#dialog & \#utt & \#pred & cross ratio\\
        \hline
        DuConv & 3,000 & 27,198 & 33,673 & 21.89\% \\
        NewsDialog & 200 & 6,037 & 3,621 & 20.01\%  \\
        PersonalDialog & 300 & 1,579 & 1,441 & 12.56\% \\
    \bottomrule
    \end{tabular}
    \egroup
    \caption{Statistics of the annotations on DuConv, NewsDialog and PersonalDialog.}
    \label{tab:statistics}
\end{table}
\section{Statistics of the datasets}
Table~\ref{tab:statistics} shows the statistics of three datasets, where \textit{cross ratio} is the percent of
arguments that are in different utterances with the predicate.

\end{CJK*}
\end{document}